\RequirePackage{lineno}
\documentclass{elsarticle}

\usepackage{amssymb}
\usepackage{graphicx}
\usepackage{multirow}
\usepackage{subfig}
\usepackage{setspace}
\usepackage{hyperref}
\usepackage{url}
\usepackage{amsmath}
\begin{document}
%\linenumbers
\title{Generalized Canonical Correlation Analysis for Disparate Data Fusion}
%\date{March 14, 2012}

\author[msun]{Ming~Sun}
\ead{msun8@jhu.edu}

\author[cep]{Carey~E.~Priebe\corref{cor1}}
\ead{cep@jhu.edu}

\author[mtang]{Minh~Tang}
\ead{mtang10@jhu.edu}

\cortext[cor1]{Corresponding author. Phone: 410-516-7200. Fax: 410-516-7459.}
\address[msun]{Department of Electrical and Computer Engineering, Johns Hopkins University, Baltimore, MD 21218, USA.}
\address[cep]{Department of Applied Mathematics and Statistics, Johns Hopkins University, Baltimore, MD 21218, USA.}
\address[mtang]{Department of Applied Mathematics and Statistics, Johns Hopkins University, Baltimore, MD 21218, USA.}

\markboth{}{}

\begin{abstract}
%\boldmath
Manifold matching works to identify embeddings of multiple disparate data spaces into the same low-dimensional space, where joint inference can be pursued. It is an enabling methodology for fusion and inference from multiple and massive disparate data sources. In this paper we focus on a method called Canonical Correlation Analysis (CCA) and its generalization Generalized Canonical Correlation Analysis (GCCA), which belong to the more general Reduced Rank Regression (RRR) framework. We present an efficiency investigation of CCA and GCCA under different training conditions for a particular text document classification task.
\end{abstract}


\maketitle

\section{Introduction}

\subsection{Purpose}
In the real world, one single object may have different representations in different domains. For example, the Declaration of Independence has versions translated into different languages. Let $n$ denote the number of objects $O_i,~i = 1, \ldots, n$, and $K$ be the number of domains. Then we have
\begin{equation}
\mathbf{x}_{i1}\sim\cdots\sim\mathbf{x}_{ik}\sim\cdots\sim\mathbf{x}_{iK},~i = 1, \ldots, n
\end{equation}
where the $i$th object $O_i$ has $K$ measurements $\mathbf{x}_{ik},~k = 1, \ldots, K$; $\mathbf{x}_{ik}\in\Xi_{k}$ is the representation for object $O_i$ in space $\Xi_k$.

\begin{figure}[!t]

\centering
\includegraphics[width=3.5in]{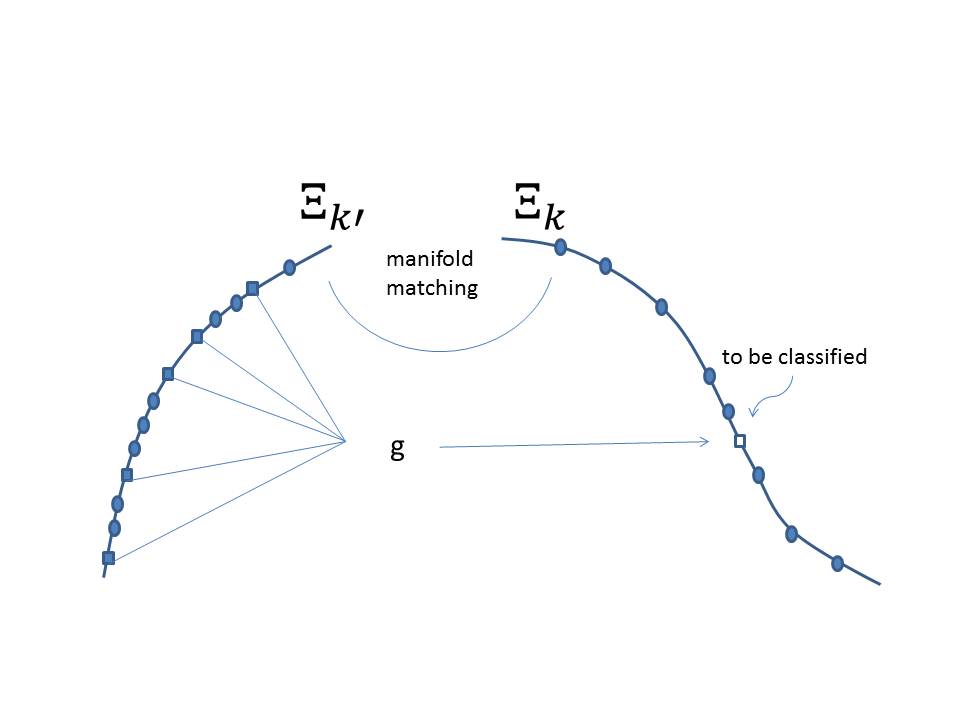}
\caption{Classification problem}
\label{classification}
\end{figure}

The problem explored in this paper is that for $m$ new objects $O'_i,~i = 1, \ldots, m$, how to classify their representations $\mathbf{y}_{ik}\in\Xi_k$ given the representations $\mathbf{y}_{i'k'}\in\Xi_{k'}$ with $k \neq k'$. For this task, $\mathbf{x}_{ik},~\mathbf{x}_{ik'},~i = 1, \ldots, n$, described above are needed to learn the relation between $\Xi_k$ and $\Xi_{k'}$ so that we can map data from $\Xi_k$ and $\Xi_{k'}$ to a common space $\chi$. Thus $\mathbf{x}_{ik},~\mathbf{x}_{ik'}$ are the domain relation learning training data. In our scenario, we are interested in a particular setting that the data to be classified is in separated classes different from the data used to learn the low dimensional manifold. This is shown in Figure \ref{classification}, where disks represent the domain relation learning training data $\mathbf{x}_{ik},~\mathbf{x}_{ik'}$ and squares denote the classifier training and testing data $\mathbf{y}_{ik}, \mathbf{y}_{i'k'}$. A classification rule $g$ is trained on $\mathbf{y}_{i'k'}$ and applied on $\mathbf{y}_{ik}$. We consider one domain relation learning method, Canonical Correlation Analysis (CCA) \cite{hotelling36, hardoon04}, which can be carried out using reduced-rank regression routines \cite{izenman75, izenman08}. We investigate classification performance in the common space $\chi$ obtained via CCA, training the classifier on $\mathbf{y}_{i'k'}$ and testing on $\mathbf{y}_{ik}$. The focus of this paper is not on optimizing the classifier; rather, we investigate performance for a given clasifier (5-Nearest Neighbor) as a function of the number of domain relation learning training data observation $n$ used to learn $\chi$. The main contribution of this paper is an investigation of the notion of supplementing the training data of classifier by using data from other disparate sources/spaces.

\subsection{Summary}
The structure of the paper is as follows: Section \ref{background} talks about related work. Section \ref{method} discusses the methods employed, including the manifold matching framework as well as embedding and classification details. Experimental setup and results are presented in section \ref{results}. Section \ref{conclusion} is the conclusion.

\section{Background}\label{background}
Different methods of transfer learning, multitask learning and domain adaptation are discussed in a recent survey \cite{pan10}. There are algorithms developed on unsupervised document clustering where training and testing data are of different kinds \cite{karakos07}. The problem explored in this paper can be viewed as a domain adaptation problem, for which the training and testing data of the classifier are from different domains. When the classification is on the text documents in different languages, as described in the later sections of this paper, it is called cross-language text classification. There is much work on inducing correspondences between different language pairs, including using bilingual dictionaries \cite{olsson05}, latent semantic analysis (LSA) features \cite{dumais97}, kernel canonical correlation analysis (KCCA) \cite{li07}, etc. Machine translation is also involved in the cross-language text classification, which translates the documents into a single domain \cite{rigutini05, fortuna05, ling08}.

\section{Method}\label{method}
In this paper, we focus on manifold matching. The whole procedure can be divided into the following steps:
\begin{enumerate}
\item[$\bullet$] For each single space $\Xi_k$, calculate the dissimilarity matrix for all domain relation learning training data observations $O_i$.
\item[$\bullet$] For each $k$, use Multidimensional Scaling (MDS) on the dissimilarity matrix to get a Euclidean representation $E_k$.
\item[$\bullet$] Run CCA (for $K=2$) or Generalized CCA ($K>2$) to map the collection $E_1, \ldots, E_K$ to a common space $\chi$.
\item[$\bullet$] Pursue joint inference (i.e. classification) in the common space $\chi$.
\end{enumerate}

This procedure combines MDS and (Generalized) CCA in a sequential way. Firstly MDS is applied to learn low-dimensional manifolds, then (Generalized) CCA is used to match those manifolds to obtain a common space.

This paper focuses on manifold matching and it demonstrates the classification improvement via fusing data from additional space to learn the common low dimensional manifold. It is interesting to investigate how to generate the low dimensional space using all data instead of matching separate manifolds. But this requires calculating the dissimilarity information for the objects' representation in different spaces properly for the multi-dimensional scaling purpose. This issue had been investigated, e.g., \cite{ma12, cep12}, but there had not been any clear answer.

\subsection{Manifold Matching Framework}
\begin{figure}[!t]

\centering
\includegraphics[width=3.5in]{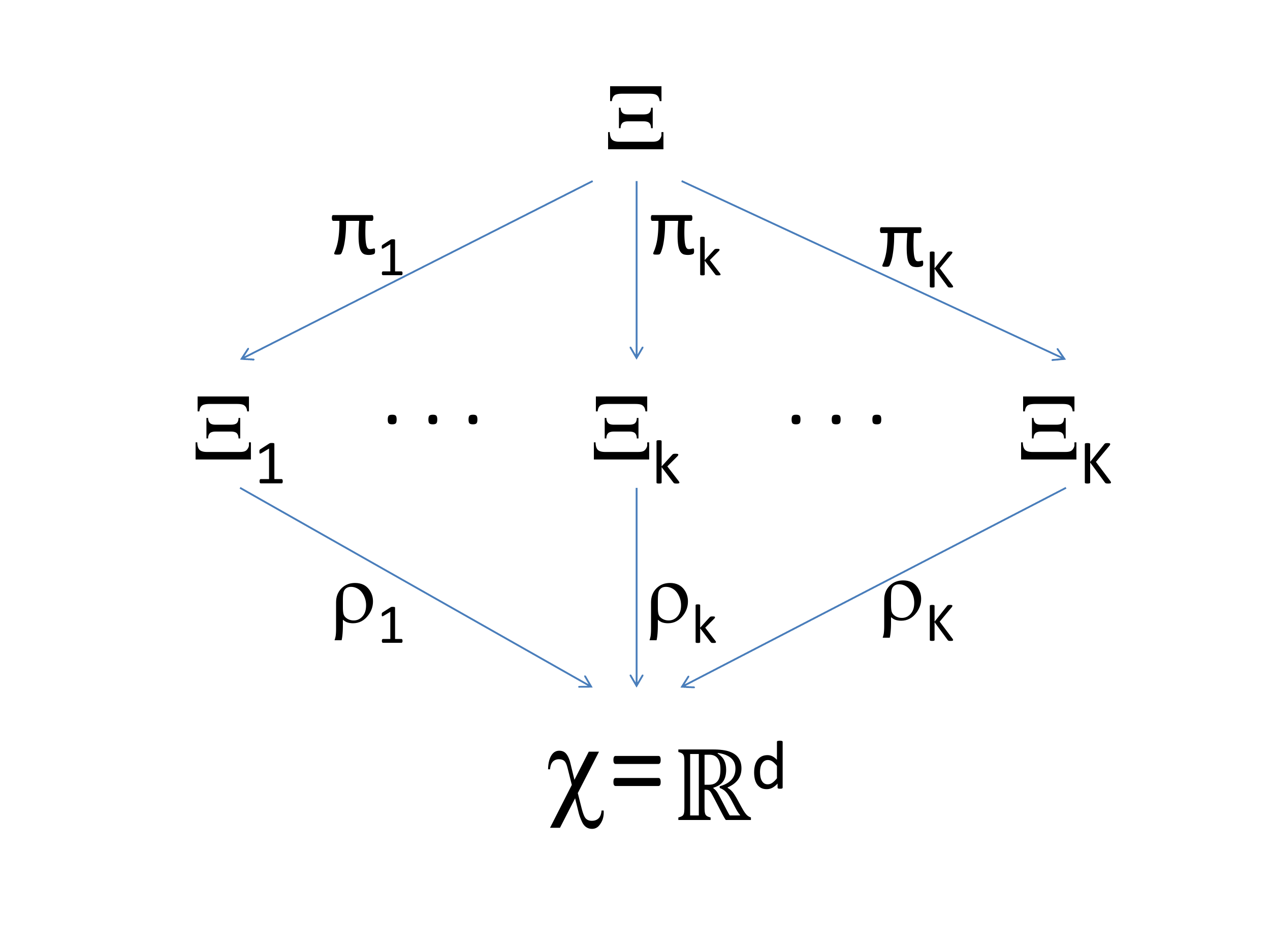}
\caption{Manifold matching model}
\label{model}
\end{figure}

The framework structure for manifold matching is shown in Figure \ref{model} \cite{ma12, cep12}. 
For each of the $n$ objects $O_i \in \Xi,~i = 1, \ldots, n$, there are $K$ representations $\mathbf{x}_{ik} \in \Xi_k,~k = 1, \ldots, K$ generated by the mappings $\pi_k$. Manifold matching works to find $\rho_1, \ldots, \rho_K$ to map $\mathbf{x}_{i1}, \ldots, \mathbf{x}_{iK}$ to a low-dimensional common space $\chi = \mathbb{R}^d$:
\begin{equation}
\tilde{\mathbf{x}}_{ik} = \rho_{k}(\mathbf{x}_{ik}),~i = 1, \ldots, n,~k = 1, \ldots, K.
\end{equation}

After learning the $\rho_k$s, we can map a new measurement $\mathbf{y}_k \in \Xi_{k}$ into the common space $\chi = \mathbb{R}^d$ via:
\begin{equation}
\tilde{\mathbf{y}}_{k} = \rho_{k}(\mathbf{y}_{k})
\end{equation}
This allows joint inference to proceed in $\mathbb{R}^d$.

\subsection{Embedding} \label{embedding}
The work described in this paper is based on dissimilarity measures. Let $\delta_k$ denote the dissimilarity measure in the $k$th space $\Xi_k$, and $\tilde{\delta}$ be the Euclidean distance in the common space $\mathbb{R}^d$. There are two kinds of mapping errors induced by the $\rho_k$s: fidelity error and commensurability error.

Fidelity measures how well the original dissimilarities are preserved in the mapping $\mathbf{x}_{ik}\mapsto \tilde{\mathbf{x}}_{ik}$, and the fidelity error is defined as the within-condition squared error:
\begin{equation}
\epsilon_{f_k}^{2} = \frac{1}{\binom{n}{2}}\sum_{1\leq i<j\leq n}(\tilde{\delta}(\tilde{\mathbf{x}}_{ik}, \tilde{\mathbf{x}}_{jk}) - \delta_k(\mathbf{x}_{ik}, \mathbf{x}_{jk}))^2
\end{equation}

Commensurability measures how well the matchedness is preserved in the mapping, and the commensurability error is defined as the between-condition squared error:
\begin{equation}
\epsilon_{c_{k_1k_2}}^{2} = \frac{1}{n}\sum_{1\leq i \leq n}(\tilde{\delta}(\tilde{\mathbf{x}}_{ik_1}, \tilde{\mathbf{x}}_{ik_2}) 
%- \delta_{k_1k_2}(\mathbf{x}_{ik_1}, \mathbf{x}_{ik_2})
)^2
\end{equation}

%\subsubsection{Multidimensional Scaling}\label{proc}
Multidimensional Scaling (MDS) \cite{torgerson52, cox01, borg05} works to get a Euclidean representation while approximately preserving the dissimilarities. Given the $n \times n$ dissimilarity matrix $\Delta_k = [\delta_k(\mathbf{x}_{ik}, \mathbf{x}_{jk})]$ in space $\Xi_k$, multidimensional scaling generates embeddings $\tilde{\mathbf{x}}_{ik}' \in \mathbb{R}^{d'}$ for $\mathbf{x}_{ik} \in \Xi_k,~i = 1, \ldots, n,~k = 1, \ldots, K$, which attempts to optimize fidelity, that is, $||\tilde{\mathbf{x}}_{ik}'-\tilde{\mathbf{x}}_{jk}'||\approx \delta_k(\mathbf{x}_{ik}, \mathbf{x}_{jk})$.

For the $K=2$ case, multidimensional scaling generates $n \times d'$ matrices $\tilde{X}'_1$ from $\Delta_1$ and $\tilde{X}'_2$ from $\Delta_2$. The $i$th row vector $\tilde{\mathbf{x}}_{ik}'$ of  $\tilde{X}_k'$ is the multidimensional scaling embedding for $\mathbf{x}_{ik}$.

%\subsubsection{Canonical Correlation Analysis (CCA)}\label{cca}
Canonical correlation analysis is applied to the multidimensional scaling results. Canonical correlation works to find $d' \times d$ matrices $U_1:\tilde{X}_1' \mapsto \tilde{X}_1$ and $U_2:\tilde{X}_2' \mapsto \tilde{X}_2$ as the linear mapping method to maximize correlation for the mappings into $\mathbb{R}^{d}$, where two matices satisfy $U_1^T U_1=I$ and $U_2^T U_2=I$. That is, for the $l$th ($1\leq l\leq d$) dimension, the mapping process is defined by $\mathbf{u}_1^l$ and $\mathbf{u}_2^l$, the $l$th column vector of $U_1$ and $U_2$ respectively. The orthonormal requirement on the columns of $U_1$ (similarly $U_2$) implies that the correlation between different dimensions of the embedding is 0. The correlation of the mapping data is calculated as 
\begin{equation}
\rho_l = \frac{(\tilde{X}_1' \mathbf{u}_1^l)^T(\tilde{X}_2' \mathbf{u}_2^l)}{\parallel\tilde{X}_1' \mathbf{u}_1^l\parallel\parallel\tilde{X}_2' \mathbf{u}_2^l\parallel}
\end{equation}
which is equivalent to 
\begin{equation}
\rho_l = (\tilde{X}_1' \mathbf{u}_1^l)^T(\tilde{X}_2' \mathbf{u}_2^l)
\end{equation}

subject to

\begin{equation}
(\tilde{X}_1' \mathbf{u}_1^l)^T(\tilde{X}_1' \mathbf{u}_1^l) = (\tilde{X}_2' \mathbf{u}_2^l)^T(\tilde{X}_2' \mathbf{u}_2^l) = 1
\end{equation}

And the constraint can be proved to be equivalent to
\begin{equation}
\frac{(\tilde{X}_1' \mathbf{u}_1^l)^T(\tilde{X}_1' \mathbf{u}_1^l) + (\tilde{X}_2' \mathbf{u}_2^l)^T(\tilde{X}_2' \mathbf{u}_2^l)}{2} = 1
\end{equation}

For CCA it holds $\rho_1\geq\rho_2\geq\ldots\geq\rho_d$.

For new data $\mathbf{y}_{k},~k = 1, 2$, out-of-sample embedding for multidimensional scaling \cite{anderson03, trosset08} generates $d'$ dimensional row vector $\tilde{\mathbf{y}}_{k}'$. The final embeddings in the common space $\mathbb{R}^d$ are given by $\tilde{\mathbf{y}}_{1} = \tilde{\mathbf{y}}_{1}' U_1$ and $\tilde{\mathbf{y}}_{2} = \tilde{\mathbf{y}}_{2}' U_2$.

Canonical correlation analysis optimizes commensurability without regard for fidelity \cite{cep12}. For our work, first we use multidimensional scaling to generate a fidelity-inspired Euclidean representation, and then we use canonical correlation analysis to enforce low dimensional commensurability.

Canonical correlation analysis is developed as a way of measuring the correlation of two multivariate data sets, and it can be formulated as a generalized eigenvalue problem. The expansion of canonical correlation analysis to more than two multivariate data sets is also available \cite{kettenring71}, which is called Generalized Canonical Correlation  Analysis (GCCA). Generalized canonical correlation analysis simultaneously find $U_1:\tilde{X}_1' \mapsto \tilde{X}_1, \ldots, U_K:\tilde{X}_K' \mapsto \tilde{X}_K$ to map the multivariate data sets in $K$ spaces to the common space $\mathbb{R}^{d}$. Similarly for the new data $\mathbf{y}_{k},~k = 1, \ldots, K$, we can get their representations in the common space $\mathbb{R}^d$ as $\tilde{\mathbf{y}}_{1} = \tilde{\mathbf{y}}_{1}' U_1, \ldots, \tilde{\mathbf{y}}_{K} = \tilde{\mathbf{y}}_{K}' U_K$. Similar to CCA, the correlation of data in the $l$th mapping dimension is calculated as \cite{via05}

\begin{equation}
\rho_l = \frac{1}{K(K-1)}\sum_{g, h = 1}^{K}(\tilde{X}_g' \mathbf{u}_g^l)^T(\tilde{X}_h' \mathbf{u}_h^l) 
\end{equation}

subject to

\begin{equation}
\frac{1}{K}\sum_{g=1}^{K} (\tilde{X}_g' \mathbf{u}_g^l)^T(\tilde{X}_g' \mathbf{u}_g^l) = 1
\end{equation}

GCCA can be formulated as a generalized eigenvalue problem. Different algorithms have been developed as the solution, e.g. least square regression. For the particular dataset used in our experiments, because it is not very large, we can perform eigenvalue decomposition on the respective matrices directly.

\subsection{Classification}
Given the measurements of $m$ new data points $\mathbf{y}_{ik},~i = 1, \ldots, m,~k = 1, \ldots, K$, (generalized) canonical correlation analysis in section \ref{embedding} yields the embeddings $\tilde{\mathbf{y}}_{ik}$ in the common space $\mathbb{R}^{d}$. To classify $\tilde{\mathbf{y}}_{ik}$, instead of using data points from the same space $\Xi_k$ (i.e. $\tilde{\mathbf{y}}_{i'k},~i'\neq i$), we consider the problem in which we must borrow the embeddings from another space $\Xi_{k'}$ for training, that is, $\tilde{\mathbf{y}}_{i'k'},~i'\neq i,~k'\neq k$. This problem is motivated by the fact that in many situations there is a lack of training data in the space where the testing data lie.

\subsection{Efficiency Investigation}
We investigate the effect of the number of domain relation learning training data observations on the classification performance.

\section{Experiments Results}\label{results}
%\subsection{Theory?}
%\subsection{Simulation?}
\subsection{Dataset}
Our experiments apply canonical correlation analysis and its generalization to text document classification. The dataset is obtained from wikipedia, an open-source multilingual web-based encyclopedia with  around 19 million articles in more than 280 languages. Each document may have links pointing to other documents in the same language which explain certain terms in its content as well as the documents in other languages for the same subject. Articles of the same subject in different languages are not necessarily the exact translations of one another. They can be written by different people and their contents can differ significantly.

English articles within a 2-neighborhood of the English article "Algebraic Geometry" are collected. The corresponding French documents of those English ones are also collected. So this data set can be viewed as a two space case: $\Xi_1$ is the English space and $\Xi_2$ is the French space. There are in total 1382 documents in each space. That is, $\mathbf{z}_{1,1},\ldots, \mathbf{z}_{1382,1} \in \Xi_1$, and $\mathbf{z}_{1,2},\ldots,  \mathbf{z}_{1382,2} \in \Xi_2$. Note that $\mathbf{z}_{ik},~i = 1, \ldots, 1382,~k = 1, 2$ includes both domain relation learning training data $\mathbf{x}_{ik},~i = 1, \ldots, n$ and new data points $\mathbf{y}_{ik},~i = 1, \ldots, m$ ($m+n = 1382$) used for classification training and testing.

All 1382 documents are manually labeled into 5 disjoint classes ($0-4$) based on their topics. The topics are category, people, locations, date and math things respectively. There are 119 documents in class 0, 372 documents in class 1, 270 documents in class 2, 191 documents in class 3, and 430 documents in class 4. The documents in classes $0, 2, 4$ are the domain relation learning training data $\mathbf{x}_{ik},~i = 1, \ldots, n,~k = 1, 2$. There are in total $819$ documents in those 3 classes ($n = 819$). The $563$ ($m = 563$) documents in classes $1, 3$ are the new data $\mathbf{y}_{ik}, i = 1, \ldots, m, k = 1, 2$. They are used to train a classifier and run the classification test.

\subsection{Dissimilarity Matrix} \label{dismat}
The method described in section \ref{embedding} starts with the dissimilarity matrix. For our work two different kinds of dissimilarity measures are considered: text content dissimilarity matrix $\Delta_{k}^{t}$ and graph topology dissimilarity matrix $\Delta_{k}^{g}$. Both matrices are of dimension $1382 \times 1382$, containing the dissimilarity information for all data points $\mathbf{z}_{1k}, \ldots, \mathbf{z}_{1382k}$.

Graphs $G_k(V, E_k)$ can be constructed to describe the dataset; $V$ represents the set of vertices which are the 1382 wikipedia documents, and $E_k$ is the set of edges connecting those documents in language $k$.

The ($i, j$) entry $\Delta_{k}^{g}(i,j)\in \Delta_{k}^{g}$ is the number of steps on the shortest path from document $i$ to document $j$ in $G_k$. In the English space $\Xi_1$, $\Delta_{1}^{g}(i,j) \in \{0, \ldots, 4\}$, where the 4 comes from the 2-neighborhood document collection. In the French space $\Xi_2$, $\mathbf{z}_{i2} \in \Xi_{2}$ is the document in French corresponding to the document $\mathbf{z}_{i1}\in\Xi_1$, and $\Delta_{2}^{g}(i,j) \in \Delta_{2}^{g}$ depends on the French graph connections. It is possible that $\Delta_{2}^{g}(i,j) \neq \Delta_{1}^{g}(i,j)$. At the extreme end, $\Delta_{2}^{g}(i,j) = \infty$ when $\mathbf{z}_{i2}$ and $\mathbf{z}_{j2}$ are not connected. We set $\Delta_{2}^{g}(i,j) = 6$ for $\Delta_{2}^{g}(i,j) > 4$.

$\Delta_{k}^{t}(i,j) \in \Delta_{k}^{t}$ is based on the text processing features for documents $\mathbf{z}_{ik}, \mathbf{z}_{jk} \in \Xi_k$. Given the feature vectors $\mathbf{f}_{ik}, \mathbf{f}_{jk}$, $\Delta_{k}^{t}(i,j)$ is calculated by the cosine dissimilarity $\Delta_{k}^{t}(i,j) = 1 - \frac{\mathbf{f}_{ik} \cdot \mathbf{f}_{jk}}{\|\mathbf{f}_{ik}\|_2 \|\mathbf{f}_{jk}\|_2}$. For our experiments, we consider three different features for $\mathbf{f}$: mutual information (MI) features \cite{lin02, pantel02, cep04}, term frequency-inverse document frequency (TFIDF) features \cite{salton88} and latent semantic indexing (LSI) features \cite{deerwester90}. The wikipedia dataset used in the experiments are available online \footnote{\url{http://www.cis.jhu.edu/~zma/zmisi09.html}}. See the paper \cite{cep09} for more details/description.

\subsection{Embedding Dimension Selection}
\begin{figure}[!t]

\centering
\includegraphics[width=3.5in]{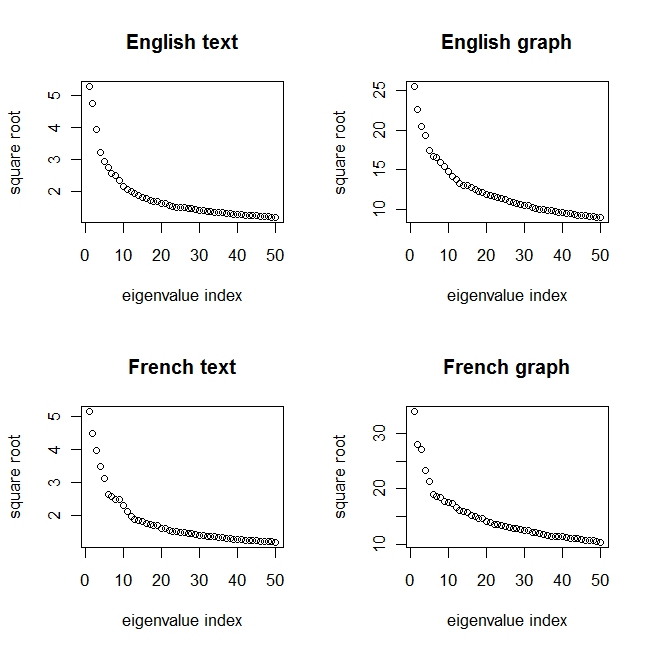}
\caption{Square root of eigenvalues for covariance matrix (all data used)}
\label{eigplot_all}
\end{figure}

To choose the dimension $d$ for the common space $\mathbb{R}^d$, we pick a sufficiently large dimension and embed $\Delta_{k}^{t}$ and $\Delta_{k}^{g}$ via multidimensional scaling. The scree plot for the MDS embedding is shown in Fig \ref{eigplot_all} (term frequency-inverse document frequency features are used for the text dissimilarity calculation).

Based on the plots in Figure \ref{eigplot_all}, we choose $d = 15$ for the dimension of the joint space $\chi$, which is low but preserves most of the variance \cite{jolliffe02}. This model selection choice of dimension is an important issue in its own right; for this paper, we fix $d = 15$ throughout.

For the canonical correlation analysis step, since it requires to multidimensional scale the dissimilarity matrices to $d'$ at the beginning, as described in section \ref{embedding}, when we choose different number $n'$ of domain relation learning training documents, $d'$ depends on $n'$. The choice of dimension is once again an important model selection problem; for this paper, the values of $d'$ with different $n'$ are shown in Table \ref{ccadim}. We believe that the values of $d'$ are chosen large enough to preserve most of the structure yet still small enough to avoid dimensions of pure noise which might deteriorate the following (G)CCA step. The second column indicates what percentage of the total manifold matching training data $\mathbf{x}_{ik}$ is used.

\begin{table}[!t]
%% increase table row spacing, adjust to taste
\renewcommand{\arraystretch}{1.3}
% if using array.sty, it might be a good idea to tweak the value of
% \extrarowheight as needed to properly center the text within the cells
\caption{MDS Dimensions}
\label{ccadim}
\centering
%% Some packages, such as MDW tools, offer better commands for making tables
%% than the plain LaTeX2e tabular which is used here.
\begin{tabular}{|c|c|c|}
\hline
$n'$ & $\%~of~n$ & $d'$ \\ 
\hline
\hline
82 & $10\%$ & 40 \\
\hline
164 & $20\%$ & 80 \\
\hline
246 & $30\%$ & 100 \\
\hline
328 & $40\%$ & 100 \\
\hline
410 & $50\%$ & 150 \\
\hline
491 & $60\%$ & 150 \\
\hline
573 & $70\%$ & 150 \\
\hline
655 & $80\%$ & 200 \\
\hline
737 & $90\%$ & 200 \\
\hline
819 & $100\%$ & 200 \\
\hline
\end{tabular}
\end{table}

\subsection{Classification Performance}
The classifier used in the experiment is $\kappa$-nearest neighbor ($\kappa$-NN). The class label of the test data is assigned by the majority class label of the $\kappa$ closest training data points. The distance used is the usual Euclidean distance. For our experiments we use the 5-nearest neighbor classifier (We do not claim that $\kappa$ = 5-NN is optimal for our experimental data. Rather, it is, illustrative; the goal of our experiments is to demonstrate the utility of using disparate domain relation learning training documents via GCCA).

There are 563 new data points $\mathbf{y}_{ik}$ in classes 1 and 3. Class 1 has 372 data points, and the remaining 191 have class label 3. For each $n'$ in Table \ref{ccadim}, we randomly sample $n'$ out of the total 819 domain relation learning training documents to learn the common space $\mathbb{R}^d$ into which we project the new data points. The classification is run in a leave-one-out way. We use 200 Monte Carlo replicates to calculate the average performance.

\begin{figure}[h!]

\centering
\subfloat[LSI Non-regularized]{
\includegraphics[width=2in]{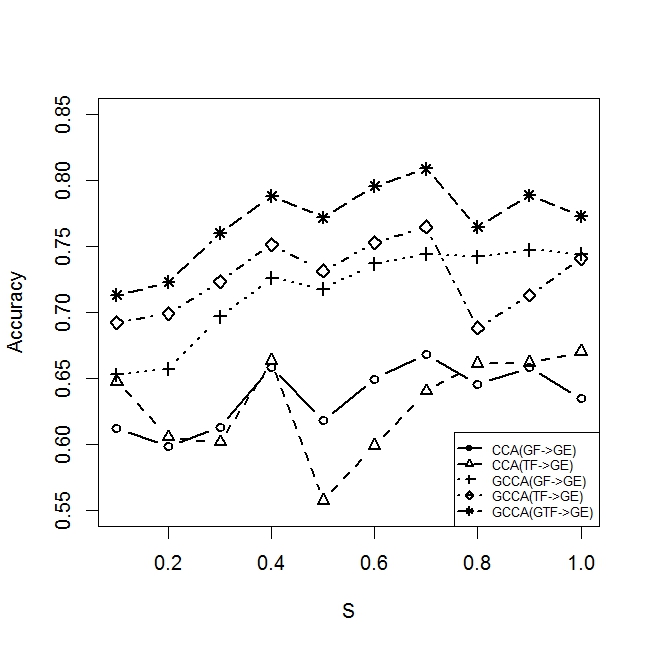}
\label{pic_selfidm_StopWordsRM_stem_lsi_CCA}
}
\hfil
\subfloat[TFIDF Non-regularized ]{
\includegraphics[width=2in]{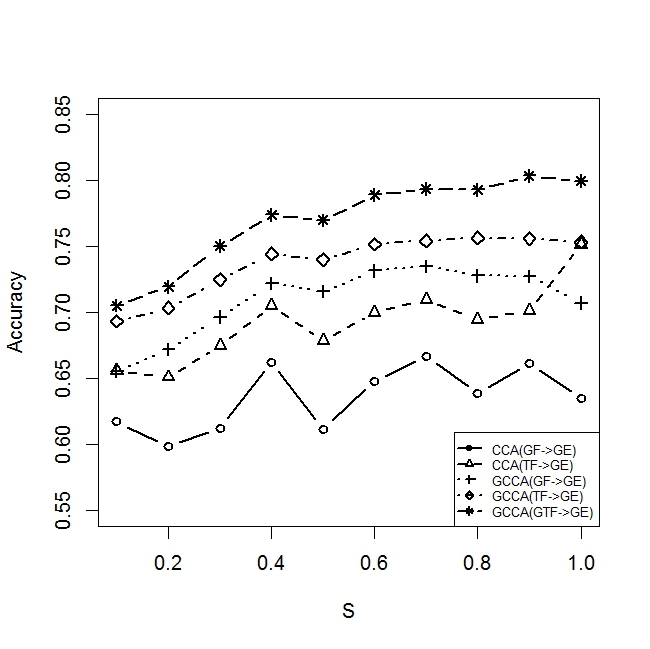}
\label{pic_selfidm_StopWordsRM_stem_tfidf_lsidim_CCA}
}
\hfil
\subfloat[MI Non-regularized]{
\includegraphics[width=2in]{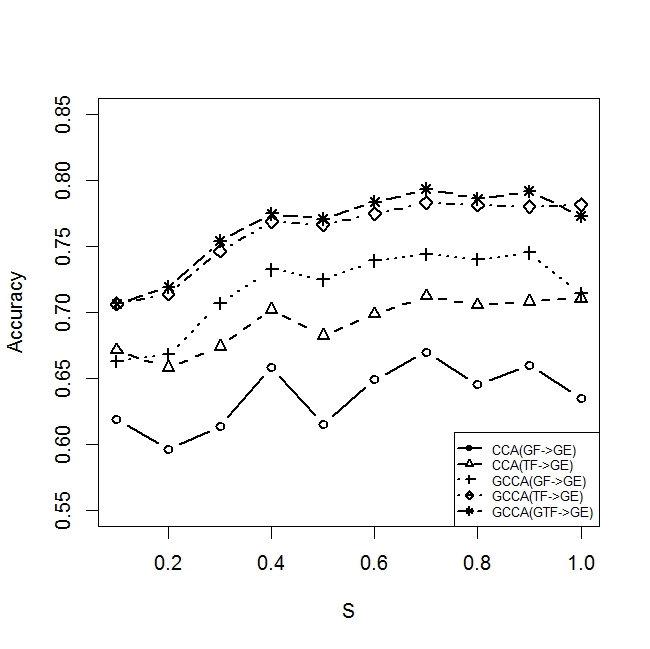}
\label{pic_selfidm_StopWordsRM_stem_mi_lsidim_CCA}}
\hfil
\subfloat[LSI Regularized]{
\includegraphics[width=2in]{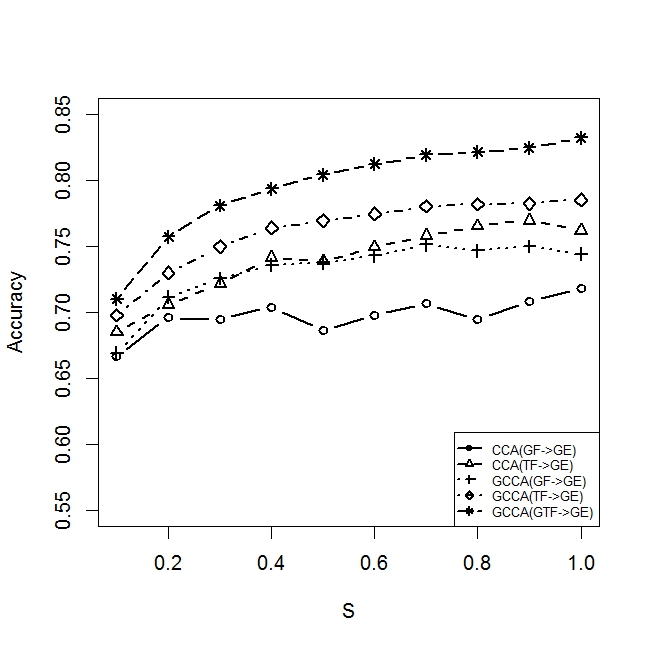}
\label{pic_selfidm_StopWordsRM_stem_lsi_halflsidim_CCA}}
\hfil
\subfloat[TFIDF Regularized]{
\includegraphics[width=2in]{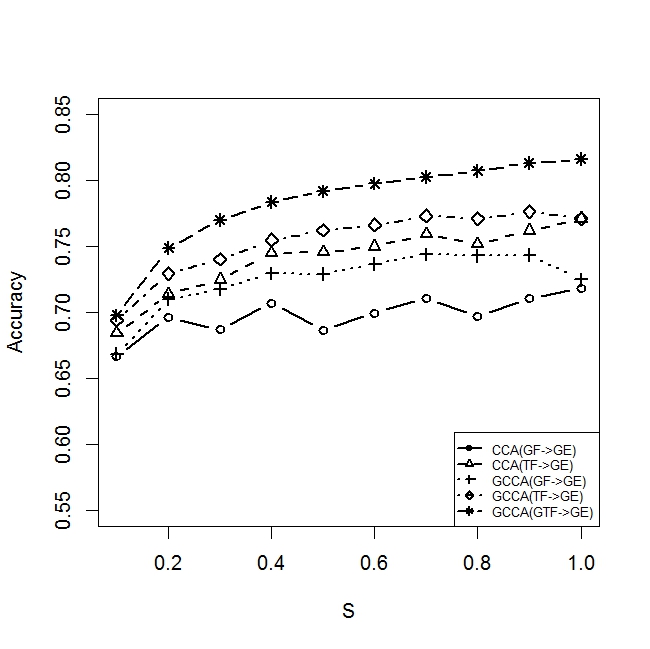}
\label{pic_selfidm_StopWordsRM_stem_tfidf_halflsidim_CCA}}
\hfil
\subfloat[MI Regularized]{
\includegraphics[width=2in]{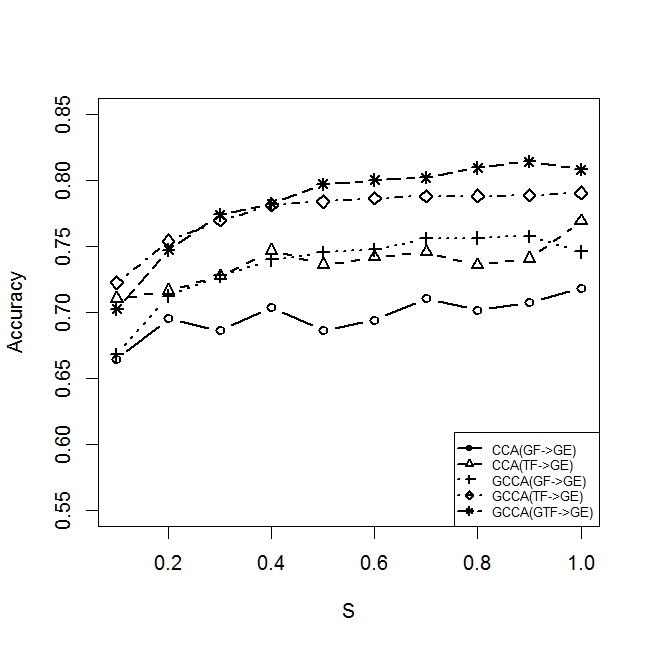}
\label{pic_selfidm_StopWordsRM_stem_mi_halflsidim_CCA}}

\caption{Classification accuracy with different amount of domain relation learning training data for (G)CCA the regularized (G)CCA with LSI, TFIDF, and MI text features}
\label{pic_CCA}
\end{figure}

The method described in section \ref{embedding} generates the embeddings $\tilde{\mathbf{y}}_{ik}\in\mathbb{R}^{15},~i = 1, \ldots, 563,~k = 1, 2$. Because there are two kinds of dissimilarity matrices considered, we have $\Delta_{k}^{t} \mapsto \tilde{\mathbf{y}}_{ik}^{t}$ and $\Delta_{k}^{g} \mapsto \tilde{\mathbf{y}}_{ik}^{g}$. The training and testing data can be chosen from not only different spaces (i.e. English space and French space), but also from different dissimilarity measures (i.e. text content dissimilarity and graph topology dissimilarity). Classification results are shown in Figures \ref{pic_selfidm_StopWordsRM_stem_lsi_CCA}, \ref{pic_selfidm_StopWordsRM_stem_tfidf_lsidim_CCA} and \ref{pic_selfidm_StopWordsRM_stem_mi_lsidim_CCA}. Note that we use different text document processing features to calculate $\Delta_{k}^{t}$. Figures \ref{pic_selfidm_StopWordsRM_stem_lsi_CCA}, \ref{pic_selfidm_StopWordsRM_stem_tfidf_lsidim_CCA} and \ref{pic_selfidm_StopWordsRM_stem_mi_lsidim_CCA} are based on the latent semantic indexing, term frequency-inverse document frequency and mutual information features respectively.

%\begin{figure}[!t]
%\caption{Classification accuracy with different amount of domain relation learning training data for (G)CCA the regularized (G)CCA with LSI, TFIDF, and MI text features}
%\centering
%\subfloat[LSI Non-regularized]{
%\includegraphics[width=2in]{pic_selfidm_StopWordsRM_stem_lsi_CCA}
%\label{pic_selfidm_StopWordsRM_stem_lsi_CCA}
%}
%\hfil
%\subfloat[TFIDF Non-regularized ]{
%\includegraphics[width=2in]{pic_selfidm_StopWordsRM_stem_tfidf_lsidim_CCA}
%\label{pic_selfidm_StopWordsRM_stem_tfidf_lsidim_CCA}
%}
%\hfil
%\subfloat[MI Non-regularized]{
%\includegraphics[width=2in]{pic_selfidm_StopWordsRM_stem_mi_lsidim_CCA}
%\label{pic_selfidm_StopWordsRM_stem_mi_lsidim_CCA}}
%\hfil
%\subfloat[LSI Regularized]{
%\includegraphics[width=2in]{pic_selfidm_StopWordsRM_stem_lsi_halflsidim_CCA}
%\label{pic_selfidm_StopWordsRM_stem_lsi_halflsidim_CCA}}
%\hfil
%\subfloat[TFIDF Regularized]{
%\includegraphics[width=2in]{pic_selfidm_StopWordsRM_stem_tfidf_halflsidim_CCA}
%\label{pic_selfidm_StopWordsRM_stem_tfidf_halflsidim_CCA}}
%\hfil
%\subfloat[MI Regularized]{
%\includegraphics[width=2in]{pic_selfidm_StopWordsRM_stem_mi_halflsidim_CCA}
%\label{pic_selfidm_StopWordsRM_stem_mi_halflsidim_CCA}}
%\end{figure}

For all three figures, the $x$-axis label $S$ indicates what proportion of the total $n$ data points are used for domain relation learning training, that is, $S = \frac{n'}{n}$; the $y$-axis is classification accuracy.

To get the solid circle curve, $\Delta_{2}^{g}$ is used for training and $\Delta_{1}^{g}$ is for testing, thus $\mathbf{x}_{ik}^{g},~i = 1, \ldots, n',~k = 1, 2$ are employed to learn the manifold matching methods. For each test data point $\tilde{\mathbf{y}}_{i1}^{g},~i \in \{1, \ldots, m\}$, the 5-NN classifier is trained on $\tilde{\mathbf{y}}_{i'2}^{g},~i' = 1, \ldots, i-1, i+1, \ldots, m$, and the classification accuracy is calculated as $m'/m$, where $m'$ is the number of correctly classified testing data points. For each $n'$, 200 Monte Carlo replicates are run to randomly sample $n'$ out of the total $n$ domain relation learning training data points $\mathbf{x}_{ik}^{g},~i = 1, \ldots, n$. The average accuracy is plotted; standard errors are available via bootstrap resampling.

The dashed triangle curve is similar to the solid circle curve except the training data is from $\Delta_{2}^{t}$ instead of $\Delta_{2}^{g}$. Since $\Delta_{2}^{t}$ and $\Delta_{1}^{g}$ are within different ranges, prescaling is needed, which is done via $\Delta_{2}^{t} = \Delta_{2}^{t}\frac{\|\Delta_{1}^{g}\|_F}{\|\Delta_{2}^{t}\|_F}$.

The remaining three curves (dotted plus, dotdash diamond, longdash asterisk) show the results of the generalized CCA, which embeds $\Delta_{1}^{g}$, $\Delta_{2}^{g}$ and $\Delta_{2}^{t}$ simultaneously to get $\tilde{\mathbf{y}}_{i1}^{g*}, \tilde{\mathbf{y}}_{i2}^{g*}$ and $\tilde{\mathbf{y}}_{i2}^{t*}~i = 1, \ldots, 563$ (with prescaling for $\Delta_{2}^{t}$ via $\Delta_{2}^{t}\frac{\|\Delta_{1}^{g}\|_F}{\|\Delta_{2}^{t}\|_F}$). For all three curves, $\tilde{\mathbf{y}}_{i1}^{g*}$ is the testing data. For the dotted plus curve, the 5-NN classifier is trained on $\tilde{\mathbf{y}}_{i2}^{g*}$. For the dotdash diamond curve, training data is $\tilde{\mathbf{y}}_{i2}^{t*}$. And the longdash asterisk curve is for the classification performance trained on $(\tilde{\mathbf{y}}_{i2}^{g*} + \tilde{\mathbf{y}}_{i2}^{t*}) / 2$.

%\newpage

Based on the results shown in Figures \ref{pic_selfidm_StopWordsRM_stem_lsi_CCA}, \ref{pic_selfidm_StopWordsRM_stem_tfidf_lsidim_CCA} and \ref{pic_selfidm_StopWordsRM_stem_mi_lsidim_CCA}, when canonical correlation analysis is used to embed the pair ($\Delta_{1}^{g}$, $\Delta_{2}^{g}$) or ($\Delta_{1}^{g}$, $\Delta_{2}^{t}$) in the same low dimensional space $\mathbb{R}^d$, $\Delta_{2}^{t}$ outperforms $\Delta_{2}^{g}$ in terms of classifying $\Delta_{1}^{g}$ for TFIDF and MI text features. But if we consider the generalized canonical correlation analysis on mapping $\Delta_{1}^{g}$, $\Delta_{2}^{g}$ and $\Delta_{2}^{t}$ to $\mathbb{R}^d$ simultaneously, it improves the embedding training in terms of classification performance. That is, to classify the embeddings of $\Delta_{1}^{g}$, the 5-NN classifer trained on $\tilde{\mathbf{y}}_{i2}^{g*}$ and tested on $\tilde{\mathbf{y}}_{i1}^{g*}$ (dotted plus curve) outperforms the one trained on $\tilde{\mathbf{y}}_{i2}^{g}$ and tested on $\tilde{\mathbf{y}}_{i1}^{g}$ (solid circle curve), and similar result holds for the pair of $\Delta_{1}^{g}$ and $\Delta_{2}^{t}$ (dotdash diamond and dashed triangle curves). This indicates incorporating information from an additional domain improves upon the embedding obtained via canonical correlation analysis in terms of classification task. The best classification results (longdash asterisk curves) come from the case not only using generalized canonical correlation analysis for domain relation learning training, but also using both $\tilde{\mathbf{y}}_{i2}^{g*}$ and $\tilde{\mathbf{y}}_{i2}^{t*}$ for classification training.

Instead of using the MDS dimensions given in Table \ref{ccadim}, we also consider a lower dimension $d''$ for each considered $n'$. By reducing the MDS dimensions, we impose additional regularization. Thus we refer to this as Regularized CCA and GCCA. How to choose the values of $d''$ properly is a non-trivial model selection problem. The values of $d''$ imply the regularization level. We choose $d''$ to be smaller than $d'$ to remove noisy dimensions from MDS embedding, but not too small to keep the fidelity of MDS. We use $d'' = d'/2$. The classification results are shown in Figures \ref{pic_selfidm_StopWordsRM_stem_lsi_halflsidim_CCA}, \ref{pic_selfidm_StopWordsRM_stem_tfidf_halflsidim_CCA} and \ref{pic_selfidm_StopWordsRM_stem_mi_halflsidim_CCA}, and they are better than the non-regularized CCA and GCCA results in Figures \ref{pic_selfidm_StopWordsRM_stem_lsi_CCA}, \ref{pic_selfidm_StopWordsRM_stem_tfidf_lsidim_CCA} and \ref{pic_selfidm_StopWordsRM_stem_mi_lsidim_CCA}, which is consistent with our expectation. The improvement of regularized CCA and regularized GCCA over their non-regularized counterparts comes from the removal of the noisy dimensions in the MDS embedding.

%\begin{figure}[!t]
%\caption{Classification accuracy with different amount of domain relation learning training data for regularized (G)CCA with LSI, TFIDF and MI text features}
%\centering
%\subfloat[LSI]{
%\includegraphics[width=2in]{pic_selfidm_StopWordsRM_stem_lsi_halflsidim_CCA}
%\label{pic_selfidm_StopWordsRM_stem_lsi_halflsidim_CCA}}
%\hfil
%\subfloat[TFIDF]{
%\includegraphics[width=2in]{pic_selfidm_StopWordsRM_stem_tfidf_halflsidim_CCA}
%\label{pic_selfidm_StopWordsRM_stem_tfidf_halflsidim_CCA}}
%\hfil
%\subfloat[MI]{
%\includegraphics[width=2in]{pic_selfidm_StopWordsRM_stem_mi_halflsidim_CCA}
%\label{pic_selfidm_StopWordsRM_stem_mi_halflsidim_CCA}}
%\end{figure}

Table \ref{clsfacc} shows the classification accuracy of various methods for $S = 10\%$ and $S = 100\%$. 

\begin{table*}[!t]
%% increase table row spacing, adjust to taste
\renewcommand{\arraystretch}{1.3}
% if using array.sty, it might be a good idea to tweak the value of
% \extrarowheight as needed to properly center the text within the cells
\caption{Classification Accuracy}
\label{clsfacc}
\centering
%% Some packages, such as MDW tools, offer better commands for making tables
%% than the plain LaTeX2e tabular which is used here.
\begin{tabular}{|c|c||c|c|c|c|}
\hline
\multicolumn{2}{|c||}{} & \multicolumn{2}{|c|}{Non-regularized} & \multicolumn{2}{|c|}{Regularized} \\
\hline
\multicolumn{2}{|c||}{} & $S=10\%$ & $S = 100\%$  & $S=10\%$ & $S = 100\%$\\
\multicolumn{2}{|c||}{} & $d' = 40$ & $d' = 200$ & $d'' = 20$ & $d'' = 100$ \\
\hline
\hline

CCA & LSI & $61.24\%\pm0.12\%$ & $63.50\%\pm0.10\%$ & $66.67\%\pm0.13\%$ & $71.84\%\pm0.10\%$\\
\multirow{2}{*}{(GF $\rightarrow$ GE)}& TFIDF & $61.73\%\pm0.12\%$ & $63.48\%\pm0.10\%$ & $66.69\%\pm0.13\%$ & $71.85\%\pm0.10\%$\\
& MI & $61.88\%\pm0.12\%$ & $63.51\%\pm0.10\%$ & $66.47\%\pm0.13\%$ & $71.85\%\pm0.10\%$\\
\hline
CCA & LSI & $64.75\%\pm0.11\%$ & $67.05\%\pm0.18\%$ & $68.51\%\pm0.14\%$ & $76.20\%\pm0.11\%$\\
\multirow{2}{*}{(TF $\rightarrow$ GE)} & TFIDF & $65.64\%\pm0.12\%$ & $75.13\%\pm0.10\%$ & $68.43\%\pm0.15\%$ & $77.09\%\pm0.11\%$\\
 & MI & $67.14\%\pm0.09\%$ & $71.05\%\pm0.09\%$ & $71.03\%\pm0.12\%$ & $76.91\%\pm0.11\%$\\
\hline
GCCA & LSI & $65.30\%\pm0.14\%$ & $74.42\%\pm0.10\%$ & $66.91\%\pm0.14\%$ & $74.42\%\pm0.08\%$\\
\multirow{2}{*}{(GF $\rightarrow$ GE)}  & TFIDF & $65.57\%\pm0.15\%$ & $70.70\%\pm0.11\%$ & $66.84\%\pm0.14\%$ & $72.47\%\pm0.09\%$\\
 & MI & $66.30\%\pm0.14\%$ & $71.40\%\pm0.10\%$ & $66.80\%\pm0.14\%$ & $74.60\%\pm0.08\%$\\
\hline
GCCA & LSI & $69.21\%\pm0.12\%$ & $74.07\%\pm0.13\%$ & $69.77\%\pm0.13\%$ & $78.51\%\pm0.07\%$\\
\multirow{2}{*}{(TF $\rightarrow$ GE)} & TFIDF & $69.33\%\pm0.13\%$ & $75.31\%\pm0.10\%$ & $69.41\%\pm0.15\%$ & $77.09\%\pm0.09\%$\\
 & MI & $70.63\%\pm0.12\%$ & $78.15\%\pm0.08\%$ & $72.24\%\pm0.12\%$ & $79.04\%\pm0.06\%$\\
\hline
GCCA & LSI & $71.31\%\pm0.11\%$ & $77.26\%\pm0.11\%$ & $71.02\%\pm0.12\%$ & $83.21\%\pm0.06\%$\\
\multirow{2}{*}{(GTF $\rightarrow$ GE)} & TFIDF & $70.53\%\pm0.11\%$ & $79.93\%\pm0.10\%$ & $69.77\%\pm0.13\%$ & $81.61\%\pm0.08\%$\\
 & MI & $70.66\%\pm0.11\%$ & $77.26\%\pm0.09\%$ & $70.23\%\pm0.12\%$ & $80.82\%\pm0.08\%$\\
\hline
\end{tabular}
\end{table*}

In the experimental settings described above, the documents in classes 1 and 3 are used for classifier training and testing, while the documents in the remaining three classes (0, 2, 4) are the domain relation learning training data. Experimental results in Figure \ref{pic_CCA} and Table \ref{clsfacc} show that GCCA is superior to CCA. However, it remains questionable whether this phenomenon holds in other settings. We investigate this problem via choosing different classes combinations for classifier training and testing. In addition to the choice of classes 1 and 3 used above, we also considered other possible combinations of two classes for classifier traning and testing. Regularized GCCA is considered here because it yields the best classification performance in the previous experimental settings. Given two classes for classifier training and testing, we use all domain relation learning training data available, that is, all the documents in the remaining three classes ($S = 100\%$). The embedding dimension for MDS is $d'' = 100$ as specified earlier in Table \ref{clsfacc}. Regarding the text feature, latent semantic indexing is selected. The results of the investigation outlined above are shown in Table \ref{allcomb}, where each row corresponds to one pair of classes. For example, the first row in Table \ref{allcomb} is for the case where classes 0 and 1 are used for classifier training and testing, and all documents in classes 2, 3, 4 are the domain relation learning training data. The results in Table \ref{allcomb} indicate that GCCA performs better than CCA for different choices of class combinations, thus strongly supporting the conclusion that GCCA is superior to CCA in terms of classification accuracy. 

\begin{table*}[!t]
%% increase table row spacing, adjust to taste
\renewcommand{\arraystretch}{1.3}
% if using array.sty, it might be a good idea to tweak the value of
% \extrarowheight as needed to properly center the text within the cells
\caption{Classification Accuracy for All Classes Combinations}
\label{allcomb}
\centering
%% Some packages, such as MDW tools, offer better commands for making tables
%% than the plain LaTeX2e tabular which is used here.
\begin{tabular}{|c||c|c|c|c|c|}
\hline
& \multicolumn{5}{|c|}{Regularized (LSI text feature), $S = 100\%$, $d'' = 100$}\\
\hline
Classification & CCA  & CCA & GCCA & GCCA & GCCA \\
Classes & (GF $\rightarrow$ GE) & (TF $\rightarrow$ GE) & (GF $\rightarrow$ GE) & (TF $\rightarrow$ GE) & (GTF $\rightarrow$ GE) \\
\hline
\hline
0, 1 & $75.36\%\pm0.04\%$ & $67.82\%\pm0.11\%$ & $80.24\%\pm0.03\%$ & $73.93\%\pm0.03\%$  & $77.39\%\pm0.04\%$\\
\hline
0, 2 & $74.29\%\pm0.06\%$ & $66.58\%\pm0.11\%$ & $83.03\%\pm0.05\%$ & $75.84\%\pm0.04\%$  & $86.89\%\pm0.05\%$\\
\hline
0, 3 & $80.00\%\pm0.08\%$ & $71.94\%\pm0.17\%$ & $85.48\%\pm0.05\%$ & $87.42\%\pm0.07\%$  & $95.81\%\pm0.04\%$\\
\hline
0, 4 & $76.14\%\pm0.05\%$ & $67.40\%\pm0.07\%$ & $78.51\%\pm0.04\%$ & $75.41\%\pm0.03\%$  & $77.41\%\pm0.04\%$\\
\hline
1, 2 & $59.19\%\pm0.07\%$ & $58.10\%\pm0.09\%$ & $61.99\%\pm0.07\%$ & $63.71\%\pm0.07\%$  & $66.98\%\pm0.06\%$\\
\hline
1, 3 & $71.84\%\pm0.10\%$ & $76.20\%\pm0.11\%$ & $74.42\%\pm0.08\%$ & $78.51\%\pm0.07\%$  & $83.21\%\pm0.06\%$\\
\hline
1, 4 & $55.74\%\pm0.06\%$ & $53.12\%\pm0.07\%$ & $61.60\%\pm0.06\%$ & $57.11\%\pm0.08\%$  & $65.84\%\pm0.06\%$\\
\hline
2, 3 & $59.22\%\pm0.12\%$ & $67.46\%\pm0.12\%$ & $64.64\%\pm0.11\%$ & $67.25\%\pm0.09\%$  & $69.85\%\pm0.09\%$\\
\hline
2, 4 & $65.71\%\pm0.07\%$ & $64.29\%\pm0.07\%$ & $71.43\%\pm0.05\%$ & $69.43\%\pm0.05\%$  & $73.00\%\pm0.04\%$\\
\hline
3, 4 & $73.11\%\pm0.08\%$ & $73.91\%\pm0.09\%$ & $76.81\%\pm0.05\%$ & $76.97\%\pm0.07\%$  & $82.13\%\pm0.05\%$\\
\hline
\end{tabular}
\end{table*}

Inferences regarding differences in the relative performance between competing methodologies (as well as the seemingly non-monotonic performance across $S$ for a given methodology) are clouded by the variability inherent in our performance estimates. However, these real-data experimental results nonetheless illustrate the general relative performance characteristics of CCA and GCCA and their regularized versions, as a function of $S$.

\section{Conclusion}\label{conclusion}
Canonical correlation analysis and its generalization are discussed in this paper as a manifold matching method. They can be viewed as reduced rank regression, and they are applied to a classification task on wikipedia documents. We show their performance with manifold matching training data from different domains and different dissimilarity measures, and we also investigate their efficiency by choosing different amounts of manifold matching training data. The experiment results indicate that the generalized canonical correlation analysis, which fuses data from disparate sources, improves the quality of manifold matching with regard to text document classification. Also, if we use regularized canonical correlation analysis and its generalization, we further improve performance.

Finally, increasing the amount of domain relation learning training data from $10\%$ to $100\%$ ($S$ in the Figures \ref{pic_selfidm_StopWordsRM_stem_lsi_CCA}, \ref{pic_selfidm_StopWordsRM_stem_tfidf_lsidim_CCA}, \ref{pic_selfidm_StopWordsRM_stem_mi_lsidim_CCA}, \ref{pic_selfidm_StopWordsRM_stem_lsi_halflsidim_CCA}, \ref{pic_selfidm_StopWordsRM_stem_tfidf_halflsidim_CCA} and \ref{pic_selfidm_StopWordsRM_stem_mi_halflsidim_CCA}) of the available 819 documents yield approximately $10\%$ improvement in classification performance. This improvement is independent of the amount of training data available for the classifier.


\begin{thebibliography}{1}

%\bibitem{IEEEhowto:kopka}
%H.~Kopka and P.~W. Daly, \emph{A Guide to \LaTeX}, 3rd~ed.\hskip 1em plus
%  0.5em minus 0.4em\relax Harlow, England: Addison-Wesley, 1999.

\bibitem{anderson03}
M.~J.~Anderson and J.~Robinson. \emph{Generalized discriminant analysis based on distances}.\hskip 1em plus 0.5em minus 0.4em\relax Australian $\&$ New Zealand Journal of Statistics, 45:301-318, 2003.

\bibitem{borg05}
I.~Borg and P.~Groenen. \emph{Modern multidimensional scaling: theory and applications}.\hskip 1em plus 0.5em minus 0.4em\relax Springer-Verlag, 2005.

\bibitem{cox01}
T.~Cox and M.~Cox. \emph{Multidimensional scaling}.\hskip 1em plus 0.5em minus 0.4em\relax Chapman and Hall, 2001.

\bibitem{deerwester90}
S.~Deerwester, S.~Dumais, T.~Landauer, G.~Furnas and R.~Harshman. \emph{Indexing by latent semantic analysis}. \hskip 1em plus 0.5em minus 0.4em\relax J. Amer. Soc. Info. Sci. 41, 391-407, 1990.

\bibitem{dumais97}
S.~T.~Dumais, T.~A.~Letsche, M.~L.~Littman and T.~K.~Landauer. \emph{Automatic cross-language retrieval using latent semantic indexing}. \hskip 1em plus 0.5em minus 0.4em\relax In AAAI Symposium on Cross Language Text and Speech Retrieval, 1997.

\bibitem{fortuna05}
B.~Fortuna and J.~Shawe-Taylor. \emph{The use of machine translation tools for cross-lingual text mining}. \hskip 1em plus 0.5em minus 0.4em\relax In Proceedings of the ICML Workshop on Learning with Multiple Views, 2005.

\bibitem{hardoon04}
D.~R.~Hardoon, S.~R.~Szedmak, and J.~R.~Shawe-taylor. \emph{Canonical correlation analysis: an overview with application to learning methods}.\hskip 1em plus 0.5em minus 0.4em\relax Neural Computation 16(12): 2639, 2004.

\bibitem{hotelling36}
H.~Hotelling. \emph{Relations between two sets of variates}.\hskip 1em plus 0.5em minus 0.4em\relax Biometrika, vol. 28, pp. 321-377, 1936.

\bibitem{izenman75}
A.~Izenman. \emph{Reduced-rank regression for the multivariate linear model}.\hskip 1em plus 0.5em minus 0.4em\relax Journal Multivariate Analysis, vol. 5, pp. 248 --264, 1975.

\bibitem{izenman08}
A.~Izenman. \emph{Modern multivariate statistical techniques: regression, classification, and manifold learning}.\hskip 1em plus 0.5em minus 0.4em\relax Springer, New York, 2008.

\bibitem{jolliffe02}
I.T.~Jolliffe. \emph{Principal component analysis} 2nd~ed. \hskip 1em plus 0.5em minus 0.4em\relax Springer, Berlin, 2002.

\bibitem{karakos07}
D.~Karakos, J.~Eisner, S.~Khudanpur and C.~E.~Priebe. \emph{Cross-instance tuning of unsupervised document clustering algorithms}. \hskip 1em plus 0.5em minus 0.4em\relax Proceedings of the Main Conference Human Language Technology Conference of the North American Chapter of the Association for Computational Linguistics, 2007.

\bibitem{kettenring71}
J.~R.~Kettenring. \emph{Canonical analysis of several sets of variables}.\hskip 1em plus 0.5em minus 0.4em\relax Biometrika, vol. 58, pp. 433-451, 1971.

\bibitem{li07}
Y.~Li and J.~Shawe-Taylor. \emph{Advanced learning algorithms for cross-language patent retrieval and classification}.\hskip 1em plus 0.5em minus 0.4em\relax Inf. Process. Manage., 43(5):1183-1199, 2007.

\bibitem{lin02}
D.~Lin and P.~Pantel. \emph{Concept discovery from text}.\hskip 1em plus 0.5em minus 0.4em\relax In Proceedings of the 19th international conference on computational linguistics, (Morristown, NJ, USA), pp. 1-7, Association for Computational Linguistics, 2002.

\bibitem{ling08}
X.~Ling, G.~Xue, W.~Dai, Y.~Jiang, Q.~Yang and Y.~Yu. \emph{Can chinese web pages be classified with english data source}? \hskip 1em plus 0.5em minus 0.4em\relax In Proceedings of WWW-08, pages 969-978, Beijing, 2008.

%\bibitem{mackay98}
%D.~MacKay. \emph{Introduction to Monte Carlo methods}. \hskip 1em plus 0.5em minus 0.4em\relax In M.~Jordan, editor, Learning in graphical models. MIT Press, 1998.

\bibitem{ma12}
Z.~Ma, D.~Marchette and C.~E.~Priebe. \emph{Fusion and inference from multiple data sources in a commensurate space}. \hskip 1em plus 0.5em minus 0.4em\relax Statistical Analysis and Data Mining, accepted for publication, January, 2012. 

\bibitem{olsson05}
J.~S.~Olsson, D.~W.~Oard and J.~Haji\v{c}. \emph{Cross-language text classification}. \hskip 1em plus 0.5em minus 0.4em\relax In Proceedings
of SIGIR-05, pp. 645-646, Salvador, 2005.

\bibitem{pan10}
S.~J.~Pan and Q.~Yang. \emph{A survey on transfer learning}.\hskip 1em plus
  0.5em minus 0.4em\relax IEEE Transactions on Knowledge and Data Engineering 22(10):1345-1359, 2010.

\bibitem{pantel02}
P.~Pantel and D.~Lin. \emph{Discovering word senses from text}.\hskip 1em plus 0.5em minus 0.4em\relax In Proceedings of ACM SIGKDD Conference on Knowledge Discovery and Data Mining, pp. 613-619, 2002.

\bibitem{cep04}
C.~E.~Priebe, D.~J.~Marchette, Y.~Park, E.~J.~Wegman, J.~L.~Solka, D.~A.~Socolinsky, D.~Karakos, K.~W.~Church, R.~Guglielmi, R.~R.~Coifman, D.~Lin, D.	~M.~Healy, M.~Q.~Jacobs, and A.~Tsao. \emph{Iterative denoising for cross-corpus discovery}.\hskip 1em plus
  0.5em minus 0.4em\relax in Proceedings of the 2004 Symposium on Computational Statistics (Invited Talk), Prague, August 23-27, 2004.

\bibitem{cep09}
C.~E.~Priebe, Z.~Ma, D.~Marchette, E.~Hohman and G.~Coppersmith. \emph{Fusion and inference from multiple data sources}.\hskip 1em plus 0.5em minus 0.4em\relax The 57th Session of the International Statistical Institute, Durban, August 16-22, 2009.

\bibitem{cep12}
C.~E.~Priebe, D.~J.~Marchette, Z.~Ma and S.~Adali. \emph{Manifold matching: joint optimization of fidelity and commensurability}.\hskip 1em plus
  0.5em minus 0.4em\relax Brazilian Journal of Probability and Statistics, accepted for publication, February, 2012.

\bibitem{rigutini05}
L.~Rigutini, M.~Maggini and B.~Liu. \emph{An em based training algorithm for cross-language text categorization}.\hskip 1em plus 0.5em minus 0.4em\relax In Proceddings of WI05, pages 529-535, Compi\'{e}gne, 2005.

\bibitem{salton88}
G.~Salton and C.~Buckley. \emph{Term-weighting approaches in automatic text retrieval}.\hskip 1em plus
  0.5em minus 0.4em\relax Information Processing $\&$ Management. Volume 24, Issue 5, p. 513-523, 1988.

%\bibitem{shakhnarovish05}
%G.~Shakhnarovish, T.~Darrell and P.~Indyk. \emph{Nearest-neighbor methods in learning and vision}. \hskip 1em plus 0.5em minus 0.4em\relax MIT Press, 2005.

\bibitem{torgerson52}
W.~Torgerson. \emph{Multidimensional scaling: I. theory and method}.\hskip 1em plus
  0.5em minus 0.4em\relax Psychometrika, 1952.

\bibitem{trosset08}
M.~W.~Trosset and C.~E.~Priebe. \emph{The out-of-sample problem for classical multidimensional scaling}.\hskip 1em plus
  0.5em minus 0.4em\relax Computational Statistics \& Data Analysis, 52(10):4635-4642, June 2008.

\bibitem{via05}
J.~Via, I.~Santamaria and J.~Perez. \emph{Canonical Correlation Analysis (CCA) Algorithms for Multiple Data Sets: Application to Blind SIMO Equalization}. \hskip 1em plus 0.5em minus 0.4em\relax 13th European Signal Processing Conference, Antalya, Turkey, 2005.

\end{thebibliography}
\end{document}